\documentclass[sigconf]{acmart}

\usepackage{booktabs} 

\usepackage{todonotes}

\usepackage[ruled,linesnumbered]{algorithm2e}
\usepackage{threeparttable}
\usepackage{tablefootnote}
\usepackage{subcaption}
\usepackage{float}

\setcopyright{rightsretained}

\acmDOI{10.475/123_4}

\acmISBN{123-4567-24-567/08/06}

\acmConference[KDD'18]{ACM Woodstock conference}{August 2018}{London, United Kingdom}
\acmYear{2018}
\copyrightyear{2016}

\acmArticle{4}
\acmPrice{15.00}

\editor{Jennifer B. Sartor}
\editor{Theo D'Hondt}
\editor{Wolfgang De Meuter}

\begin{document}
	\title{Scalable attribute-aware network embedding with locality}
	\author{Weiyi Liu, Zhining Liu}
	\authornote{These two authors are shared first authors for the paper. Datasets and a reference implementation of \emph{SANE} are available via an email to first authors.}
	\affiliation{%
		\institution{University of Electronic Science and Technology of China, China}
		\institution{IBM T.J.Watson Research Center, U.S.A}
	}
	\email{zhining.liu@ibm.com, weiyiliu@us.ibm.com}
	
	\author{Toyotaro Suzumura}
	\affiliation{%
		\institution{IBM T.J.Watson Research Center}
		\streetaddress{Yorktown Heights}
		\city{NY, 10598}
		\country{USA}}
	\email{suzumura@acm.org}
	
	\author{Guangmin Hu}
	\affiliation{%
		\institution{University of Electronic Science and Technology of China}
		\city{Chengdu}
		\state{Sichuan}
		\country{China}
	}
	\email{hgm@uestc.edu.cn}
	
	\renewcommand{\shortauthors}{Liu et al.}

	\begin{abstract}
	Adding attributes for nodes to network embedding helps to improve the ability of the learned joint representation to depict features from topology and attributes simultaneously.
	Recent research on the joint embedding has exhibited a promising performance on a variety of tasks by jointly embedding the two spaces.
	However, due to the indispensable requirement of globality based information, present approaches contain a flaw of in-scalability.

	Here we propose \emph{SANE}, a scalable attribute-aware network embedding algorithm with locality, to learn the joint representation from topology and attributes. By enforcing the alignment of a local linear relationship between each node and its K-nearest neighbors in topology and attribute space, the joint embedding representations are more informative comparing with a single representation from topology or attributes alone. And we argue that the locality in \emph{SANE} is the key to learning the joint representation at scale.

	By using several real-world networks from diverse domains, We demonstrate the efficacy of \emph{SANE} in performance and scalability aspect.
	Overall, for performance on label classification, SANE successfully reaches up to the highest F1-score on most datasets, and even closer to the baseline method that needs label information as extra inputs, compared with other state-of-the-art joint representation algorithms. What's more, \emph{SANE} has an up to 71.4\% performance gain compared with the single topology-based algorithm.
	For scalability, we have demonstrated the linearly time complexity of \emph{SANE}. In addition, we intuitively observe that when the network size scales to 100,000 nodes, the ``learning joint embedding'' step of \emph{SANE} only takes $\approx10$ seconds.
	\end{abstract}
	
	%
	%



	\maketitle
	
	\section{Introduction}
For a real network in nature, it is easy to find attributes for each node on the network. 
Taking a citation network as an example, we can quantitatively extract attribute information by counting occurrences of some keywords or other more effective methods \cite{joulin2016bag}.
Here, to reveal latent relationships among nodes, network embedding (NE) algorithms have been introduced to learn a latent, dense and low dimensional representations of nodes on the network, through topological information, such as node-pairs.
On the other hand, as attribute information on nodes offers more details for each node, to analysis the attributes, attribute embedding (AE) aims to find an appropriate latent, dense and low-dimensional representations to depict the node properties.
Hence, a natural question could be raised that, \emph{how to unify NE and AE so that the learned joint representation can simultaneously represent the node pair information in NE and the attribute information of the node in AE.} 

For instance, each publication on a citation network not only contains citation information (a.k.a, topology), but also has its content (a.k.a, attribute). If we can informatively learn a joint representation which unifies the vertex structure and its corresponding attributes, such combination should be more effective in the task of classifying each publication, compared with from topology or from attributes alone.
As such joint representation not only preserves topology structure, but also has the ability to embeds vertex attributes, edge attributes and other network related information as well.
Thus, with the fixed-size and representative embedding vectors, conventional vector-based machine learning algorithms can be naturally introduced to solve diverse problems of analysis on the network, such as node classification \cite{kazienko2012label, vachik_zhang_arxiv}, link prediction \cite{wang2016structural, icde-sutanay, zhang2016trust, Dave2018}, node clustering \cite{xu2007scan, chen2016incremental, Chen2017}, name disambiguation\cite{Zhang.Hasan.ea:15, Zhang.Saha.ea:14, tkde_online, Zhang.Dundar.ea:16,Zhang.Noman.ea:17,zhang2017purdue}, and visualization \cite{maaten2008visualizing, tang2015line}, etc.

A convenient way to achieve such goal is to transfer the network with nodes attributes into (weighted) adjacency matrix, and the corresponding attribute matrix of nodes. By manipulating these two matrices, one can get the needed joint representation. 
Most current methods for attributed network embedding are based on matrix factorization (most of them are using eigen-decomposition, see Section~\ref{RWorks} for details). However, these methods always contain a flaw that the space complexity is too large, thus gives us obstacles in 1).a relatively slow procedure of eigen-decomposition, and 2).to require a large enough space to store the adjacency matrix.

Alternatively, unlike matrix-related algorithms using global information of the network and its nodes attributes, we can design an objective that only focus on \textbf{locality}. 
That is, for network, we can seek the relationship among a set of local neighbors to depict the target node, rather than using the entire network. For nodes attributes, we can only focus on the a small set of nodes that locate in the vicinity of the target node. 
If so, the objective can be efficiently optimized using stochastic gradient updates, which increases linearly on both time and space. As a result, locality based algorithms inherently have scalability.

To tackle the aforementioned challenges, we propose \emph{SANE}, a locality-based attribute-aware network embedding algorithm for scalable learning the joint representation from both topology and attributes. 
We optimize a custom graph-based and attribute-based objective function using stochastic gradient decent.
As a result, \emph{SANE} returns a joint embedding representation with $d$-dimensional feature space that maximize the likelihood of preserving the needed information from both topology feature space and attributes feature space.

Our main contribution in this paper is to bring up a locality based joint embedding. By enforcing the alignment of a local linear relationship between each node and its K-nearest neighbors in topology space and attribute space, the joint embedding representations are more informative comparing with the ones extracted from single spaces, and more scalable than matrix factorization related algorithms.
We explore the label classification performance of \emph{SANE} on several real-world networks from diverse domains, such as social networks, information networks, system biology, etc. 
Compared with five related baseline methods, by feeding into attributes and topology, SANE successfully reaches up to the highest F1-score on most datasets, and even closer to the baseline method that needs label information as extra inputs.
What's more, \emph{SANE} has a up to 71.4\% performance gain compare with the single topology-based algorithm.
In addition, we have demonstrate the linearly time complexity of \emph{SANE}. Besides, as \emph{SANE} inherently supports for parallelization, we find that when the network size scales to 100,000 nodes, the core step of \emph{SANE} (a.k.a,``learning joint embedding'') only takes $\approx10$ seconds.
Overall, the main contributions can be summarized as follows:
\begin{enumerate}
	\item We propose a new way to jointly embed the information given from attribute and topology.
	\item We propose a novel algorithm \emph{SANE} that has comparable performance with several state-of-the-art methods and also the ability to deal with large scale network.
	\item We empirically evaluate \emph{SANE} for nodes classification on several real-world datasets.
\end{enumerate}

The remainder of the paper is organized as follows. In Section~\ref{RWorks}, we discuss recent work on NE, AE, and joint embedding, respectively. Section~\ref{PSM} describes the detailed methods within proposed algorithm \emph{SANE}. In Section~\ref{exp}, we present a thorough evaluation of the proposed algorithm. Finally, we conclude and describe future work in Section~\ref{Conclusion}.

\section{Related Works}\label{RWorks}
With the development of unsupervised feature learning techniques \cite{bengio2013representation}, deep learning methods proved successful in natural language processing tasks through neural language models \cite{dundar2015simplicity,zhang2016trust,zhang2017name}. These models have been used to capture the semantic and syntactic structures of human language \cite{collobert2008unified}, and even logical analogies \cite{mikolov2013distributed}, by embedding words as vectors.
As a graph can be interpreted as a kind of documents (by treating random walks as the equivalent of sentences),
DeepWalk \cite{perozzi2014deepwalk} introduced such methods into network analysis, allowing for the projection of network nodes into a vector space. To solve the scalability problem of this method when applied to real world information networks (which often contain millions of nodes), LINE \cite{tang2015line} was developed. LINE extended the uniform random walks of DeepWalk to 1st and 2nd order weighted random walks, and it can project a network with millions of vertices and billions of edges into a vector space in a few hours.

However, both methods have limitations.
As DeepWalk uses uniform random walks for searching, it cannot provide control over the explored neighborhoods. In contrast, LINE proposes a breadth-first strategy to sample nodes and optimize the likelihood independently over 1-hop and 2-hop neighbors, but it has no flexibility in exploring nodes at future depths. In order to deal with both of these limitations, node2vec \cite{grover2016node2vec} provides a flexible and controllable strategy for exploring network neighborhoods through the parameters $p$ and $q$. From a practical standpoint, node2vec is scalable and robust to perturbations. 

A lot of works are also focused on the analysis on attributed network due to the fact that attributed networks are common in real world, and network analysis can benefit a lot from the co-analysis of network topology and attribute. By considering the two kinds of information as two different views of node vertices, \cite{kumar2011co} proposed to apply co-regularized spectral clustering on the multi-view data collectively, and two co-regularization schemes, which are pair-wise co-regularization and centroid based co-regularization respectivel, are proposed to accomplish this. \cite{li2017attributed} proposed DANE to capture individual properties and correlations of topology and attribute. By enforcing consistency of the pairwise similarity in the origin space and embedded space, DANE can learn the embedding representation for each space. Meanwhile, DANE tries to maximize correlations of embeddings (or equivalently minimize their disagreements) to seek a consensus embedding. Furthermore, \cite{huang2017label} propose a label informed attributed network embedding (LANE) framework to enhanced DANE through incorporating label information. By jointly embedding topological structure, node attributes and label information into a low-dimensional representation, LANE achieves significantly better performance compared with the state-of-the-art embedding algorithms.

\section{Proposed Methods}\label{PSM}
Here we propose a new attribute-aware network embedding method, to unify the relationships among nodes in the topology space (Section~\ref{TE}) and attribute space (Section~\ref{AE}). 
In Section~\ref{ANE}, we demonstrate how to unify these two spaces together, and proposed a scalable and attribute-aware network embedding algorithm, \emph{SANE}.

\subsection{Topology Embedding}\label{TE}
Given a network $G=(V,E,A)$ with vertex set $V$, edge set $E$ and attribute matrix $A$, topology embedding learns a mapping function $f:V \to \mathcal{R}^{{d}_{T}}$ to transfer the relationships among nodes into a vector space $\mathcal{R}^{{d}_{T}}$, where ${{d}_{T}}$ is the chosen dimension of the vector space, and such space $\mathcal{R}^{{d}_{T}}$ can reveal a much denser and more representative way to indicate the particular nodes relationships of interest.
Various topology embedding methods generalize recent advancements in language modeling and unsupervised feature learning from sequences of words to graphs (See Section~\ref{RWorks} for details.) 

Here we choose node2vec \cite{grover2016node2vec} learn the latent representations from truncated second order random walks from a graph.
By introducing two hyper parameters $p$ and $q$ to control the random walks, node2vec added to DeepWalk the ability to control the homophily and structural equivalence properties of random walk samples. 

\begin{figure}[ht]
	{
		\centering
		\includegraphics[width=\linewidth]{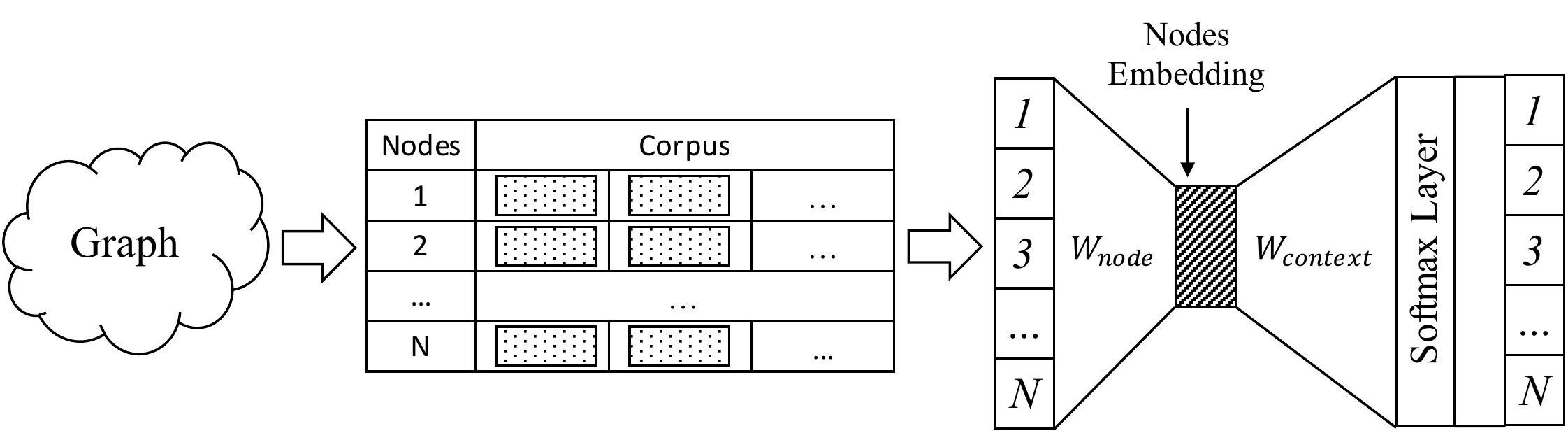}
		\caption{An overview of node2vec procedure.}
		\label{node2vec}
	}
\end{figure}

Fig.\ref{node2vec} describes the main steps for node2vec.
During the random walk procedure for each node, node2vec generates a set of fixed length nodes list to describe a target node. Based on the Skip-Gram model introduced in word2vec \cite{mikolov2013distributed}, the nodes list servers as a context to describe the target node (a.k.a corpus). 
Then, the two matrices ${{W}_{node}}$ and ${{W}_{context}}$ play the role of mapping nodes to embedding representations, and the softmax layer servers as a classifier to predicting the surrounding nodes for each node. The global objective is to maximize the Skip-Gram neural embedding model with negative sampling, defined in Eq.\ref{node2vec_eq}, using stochastic gradient updates over the generated truncated random walks $C$.

\begin{align}\label{node2vec_eq}
	\sum_{u \in V}\sum_{v \in V}\#(u,v)
	\left\{ \right. 
	\log \sigma ({\overrightarrow{W_{node}^{u}}} \cdot {\overrightarrow{W_{context}^{v}}}) + 
	\notag \\
	k\cdot {{\epsilon }_{{{v}_{N}}\sim {{P}_{C}}}}[\log \sigma (-\overrightarrow{W_{node}^{u}}\cdot \overrightarrow{W_{context}^{{{v}_{N}}}})]
	\left. \right\}
\end{align}

where $\#(u,v)$ denotes the number of times the pair $(u,v)$ appears in $C$, $\overrightarrow{W_{node}^{u}}$ denotes the embedding representation of $u$ in ${{W}_{node}}$, $k$ is the number of negative samples, ${{v}_{N}}$ is the sampled context based on the empirical unigram distribution ${{P}_{C}}(v)$, and $\sigma$ is the sigmoid funcion. Here, ${{W}_{node}}$ serves as the embedding representations of nodes on the graph for the downstream prediction task.

\subsection{Attribute Embedding}\label{AE}
Attribute matrix, of which size  is $N \times D$ ($N$ is the number of nodes in the network, and $D$ is the number of attributes for each node), is usually denser than the adjacency matrix. Embedding the nodes lying in the attribute space is more like applying dimension reduction on the attribute matrix. Various dimension reduction methods, such as principal component analysis \cite{wold1987principal}, Isomap \cite{tenenbaum2000global}, locally linear embedding \cite{roweis2000nonlinear} and so on, are proposed to extract intrinsic patterns within the data.

Here we choose the locally linear embedding (LLE) for the part of attribute embedding. LLE holds the assumption that the linear mapping between each node and its K-nearest neighbors should be preserved in the newly projected space. Let $\overrightarrow{{{A}_{i}}}\in {{\mathcal{R}}^{{{d}_{A}}}}(i=1,...,N)$ denotes the attribute for node $i$, where $d'$ is the chose dimension of the vector space for attribute embedding. LLE tries to calculate linear coefficients that reconstruct each node from its K-nearest neighbors by minimizing the reconstruction error, see Eq.\ref{LLE_W}:
\begin{equation}
\label{LLE_W}
\varepsilon (W)=\sum\limits_{i=1}^{N}{{{\left| \overrightarrow{{{A}_{i}}}-\sum\limits_{j=1}^{K}{{{W}_{ij}}\overrightarrow{{{A}_{ij}}}} \right|}^{2}}}
\end{equation}
where \emph{K} is the number of nearest neighbors that used for reconstruction, and $\overrightarrow{{{A}_{ij}}}(j=1,...,K)$ is one of the K-nearest neighbors of the node \emph{i}.
In addition, the invariance to rotations, rescalings, and translations of that data point and its neighbors are also enforced on the linear coefficients for each node \emph{i}, which leads to the constraint $\sum_{j=1}^{K}W_{ij}=1$.
The optimal weights subject to these constraints are found by solving a constrained least squares problem. 

After extracting linear relationship between each node and its K-nearest neighbors, LLE embeds all nodes in a new space while preserving the mapping as much as possible in the same time, and the error is measured by Eq.\ref{LLE_E}.
\begin{equation}
\label{LLE_E}
\varepsilon (E)=\sum\limits_{i=1}^{N}{{{\left| \overrightarrow{{{E}_{i}}}-\sum\limits_{j=1}^{K}{{{W}_{ij}}\overrightarrow{{{E}_{j}}}} \right|}^{2}}}
\end{equation}

The optimal embedding vectors can be found by the bottom \emph{d} non-zero eigen vectors of the following $N \times N$ matrix, shown in Eq.\ref{eq:eigenvector},
\begin{equation}
\label{eq:eigenvector}
{M}_{ij}={\delta}_{ij}-{W}_{ij}-{W}_{ji}+\sum_{^{k}}{{W}_{ki}{W}_{kj}}
\end{equation}
where ${\delta}_{ij}$ is 1 if $i=j$ and 0 otherwise.

\subsection{SANE: Attribute-aware Network Embedding}\label{ANE}
In this paper, we propose a new method to learn a joint embedding function $\mathcal{F}:V \bigcup A \rightarrow \mathcal{R}^{d}$ to project nodes to latent representations which not only contain information from topology but also are aware of the attributes for each node on the graph. That is, through enforcing the alignment of linear mapping from attribute and topology, the consensus embedding representations should extract useful information from two spaces simultaneously. 

Topology embedding tries to learn a latent representation for each node based on the topological features on the graph. Similarly, attribute embedding tries to project attribute vectors into a new space while preserving intrinsic patterns within attribute space. Generally speaking, regardless of whether node2vec or LLE, the goal of their mapping functions is to use the neighbors of the target node to represent the target node in the vector space. The only difference is the definition of "neighbor:" node2vec defines neighbors of a target node as a set of nodes from the truncated random walk (start from the target); while LLE treats the neighbors is a collection of nodes with similar attributes (K-nearest neighbors) to the target node. Hence, the main purpose of joint embedding function $\mathcal{F}$ proposed in this paper is by unifying the neighbor information in topological and attribute space, to learn embedding representations consistent across both spaces.

To achieve this, we enforce the alignment of the linear mapping, which extracted in the attribute space using eq.\ref{LLE_E}, between attribute space and topology space. This objective manages to maximize the degree of agreement of topology and attribute information by measuring the alignment of a linear relationship between a target node and its K-nearest neighbors. Eq.\ref{eq:Obj} shows the objective function.
\begin{align} \label{eq:Obj}
	L=\sum_{u \in V}\sum_{v \in V}\#(u,v)
	\left\{ \right.
	\log \sigma (\overrightarrow{W_{node}^{u}}\cdot \overrightarrow{W_{context}^{v}}) 
	\notag \\
	+ k\cdot {{\epsilon }_{{{v}_{N}}\sim {{P}_{C}}}}[\log \sigma (-\overrightarrow{W_{node}^{u}}\cdot \overrightarrow{W_{context}^{{{v}_{N}}}})] 
	\notag \\
	+{{|\overrightarrow{W_{node}^{u}}-\sum\limits_{j=1}^{K}{{{W}_{ij}}\overrightarrow{{W_{node}^{N_{j}^{u}}}|}^{2}}
	}}
	\left. \right\}
\end{align}
where ${N_{j}^{u}}$ is the $j$th ($j=1,...,K$) nearest neighbors of node $u$ in the attribute space.

As the embedding representations generated by word2vec (the core algorithm of node2vec) has been proven to support arithmetic calculation \cite{mikolov2013distributed}, the objective function we proposed is trying to increase the number of nodes related to arithmetic calculation from 3 (element-wise addition: $a=b-c$) to $K+1$.
We optimize a custom attribute-aware graph-based objective function using stochastic gradient updates in the same way with node2vec. Here, we do not use eigen-decomposition to optimize the objective function. Even though it loses the guarantee of the global optimum, we can still achieve acceptable performance with local optimum. Intuitively, our approach returns an optimized joint embedding that maximize the likelihood of preserving the relationships of nodes from both topology and attribute space.

In addition, Alg.\ref{Alg:AANE} gives the pseudo code for the proposed method.
Please note that, the linear mapping is only enforced on the node vectors ${W}_{node}$, rather than on ${W}_{node}$ and ${W}_{context}$ together, which is mainly for lower time complexity. In addition, the frequency of updating LLE-part loss can be reduced for further acceleration, which we have empirically proven that there is minor degradation on the performance.
\begin{algorithm}[ht]
	\caption{Attribute-aware network embedding}
	\label{Alg:AANE}
	\KwIn{$G=(V,E,A)$}
	$corpus$ $\leftarrow$ Build corpus from node2vec procedure\;
	$neighbors$, $weights$ $\leftarrow$ Build neighbors and weights from LLE procedure\;
	Optimize $\mathcal{F}$ according to Eq.\ref{eq:Obj}\;
\end{algorithm}


\section{Experiments} \label{exp}
Here we conduct various experiments to demonstrate the overall performance, parameter sensitivity, scalability, and incremental ability in exploration offered by \emph{SANE}. 

\subsection{Case Study: Karate network}\label{KN}
In Section~\ref{ANE}, we propose a method to learn a joint representation function $\mathcal{F}$ to harvest the information from topology space and attributes space simultaneously.
We now aim to empirically show that the proposed method can in fact achieve a balance in the two spaces, and to obtain a meaningful latent representation.

We use the karate network \cite{Zachary1977An}, which is a social network of friendships between 34 members of a karate club at a US university in the 1970s.
The network has 34 nodes and 77 edges. We set embedding dimension $d=2$, and run \emph{SANE} to learn the latent representation for each node in the network.
To demonstrate that \emph{SANE} can also gather useful information from attributes space, we use two kinds of communities results (i.e, ``ground truth community structure'' and ``the community structure detected by infohiermap'' ) demonstrate in \cite{cheng2014active} to assign the communities information for each node as the attributes information.
We then visualize the original network, along with the joint representations, where the colors of nodes on the diagrams indicate their attribute information (a.k.a community information).

\begin{figure*}[ht]
{
	\centering
	\resizebox{0.95\linewidth}{!}{\includegraphics[width=\linewidth]{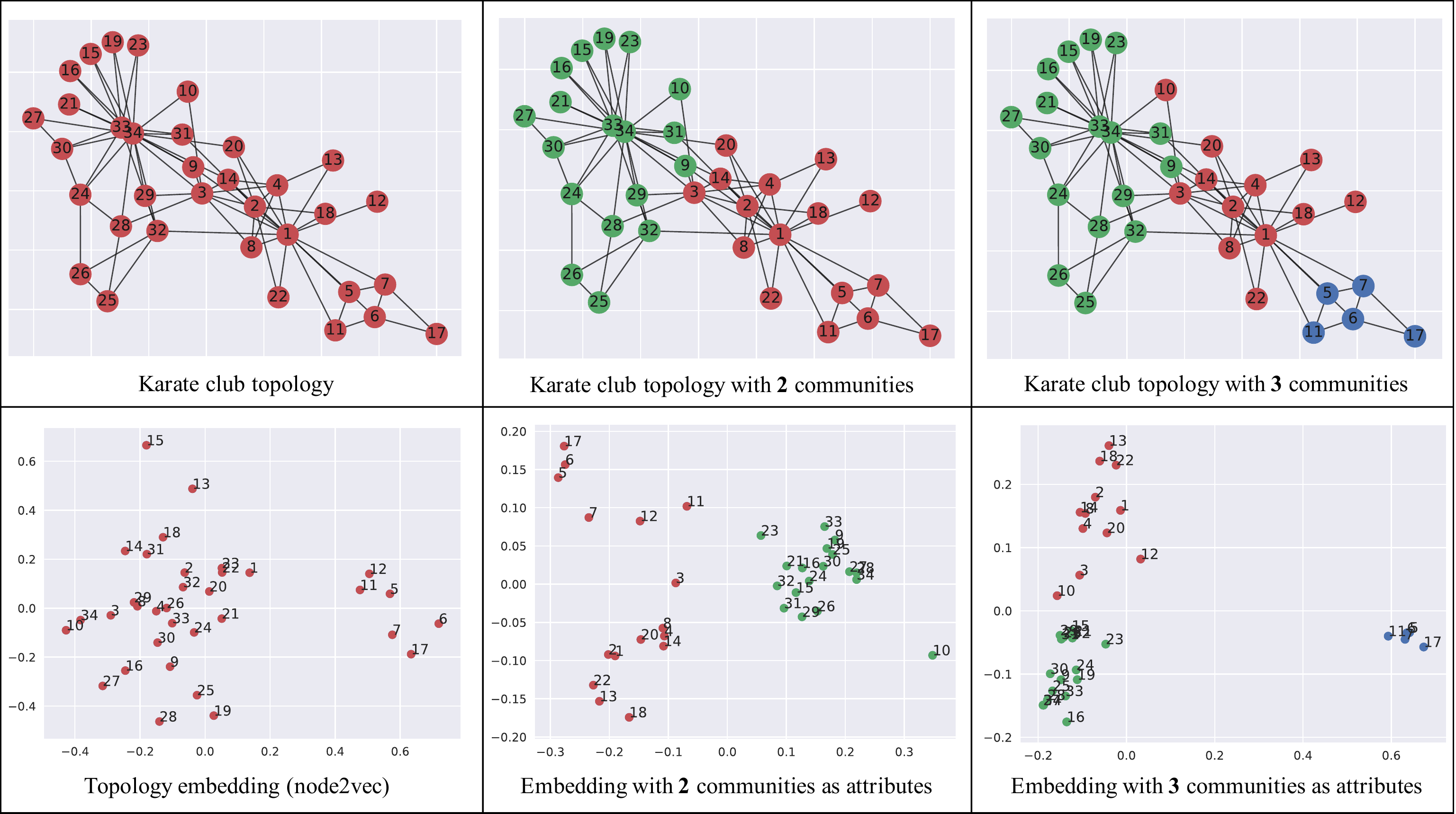}}
	\caption{Complementary visualizations of Karate club network. The first row demonstrates the topology structure of the network, along with the communities information from ``ground truth community structure (with 2 communities)'' and ``the community structure detected by infohiermap (with 3 communities.)'' The second row indicates the (joint) representations, respectively.}
	\label{fig:karate}
}
\end{figure*}

Figure~\ref{fig:karate} shows the example when we set $\lambda=1.0$ and $K=2$, and with default node2vec parameters\footnote{The default node2vec parameters are: $p=q=1$.}. Notice how attributes (i.e, network communities) affect the latent representation.
The first column showed the original node2vec topology (top) and related joint representation representation (bottom). As all nodes belong to the same community, the related representation only reflects the topological features of the graph. As can be seen from the joint representation results, this embedding can capture topological information of the karate network, and this is consistent with the nature described in the node2vec article.

In order to discover the performance of \emph{SANE} of different attributes on embedding representation results under the same topology, second and third columns use different community results as attributes information. Therefore, the resulting joint representations are also different.
First, the gain of the attribute information allows both embeddings to learn from the community information, as there is clearly linear separability between the cluster of nodes.
Second, these two embeddings ensure to some extent that the original topology information has also remained. Both embeddings can ensure that nodes with large degrees are closer in the vector space, while nodes with smaller degrees are far from the cluster centers.
For example, nodes 33, 34 and node 1, 2 are close to each other, respectively, as they are large degree nodes. While the lower degree nodes 13, 17 and 10 are distributed far away from the center.

\subsection{Dataset introduction} \label{DI}
In what follows we test our method on different attributed graphs to demonstrate its effectiveness in the joint representation of topology space and attribute space. We use the embedding vectors for label classification on nodes, and use F1-score to evaluate the performance of all methods quantitatively.

BlogCatalog and Flickr are two real-world social media datasets, which are used in \cite{huang2017label}\footnote{https://github.com/xhuang31/LANE}. Cora is a dataset based on citations between scientific papers\footnote{https://linqs.soe.ucsc.edu/data}. PPI is a graph built based the interaction between protein-protein interaction (PPI) graphs\footnote{https://downloads.thebiogrid.org/BioGRID}.
The detailed information of these datasets is shown in Table~\ref{Datasets}. Also, some summaries of each dataset are presented as follows.

\textbf{BlogCatalog} is an online blogging collaboration website. Bloggers follow each other and form a network. The number of occurrences of each keyword in bloggers' blog descriptions is set as the attribute information. The labels are selected from some predefined categories, which indicate bloggers' interests.
\textbf{Flickr} is an image and video hosting website, where users interact with each other via photo sharing. The following relationships among users are used to form a network. The attribute information is extracted from a list of tags of interest that are specified by each user. The groups that users joined are considered as labels.
\textbf{Cora} consists of 2708 scientific publications, and each of them is categorized into one of seven classes, which is considered as the labels. The network is created based the citation between these scientific publications, and every paper cites or is cited by at least one other paper. Each publication is represented as a 0/1-valued word vector indicating the absence/presence of the corresponding word from the dictionary.
\textbf{PPI} is created based on protein-protein interaction, here we try to classify protein roles concerning topology and attribute information. The attribute information is extracted from positional gene sets, motif gene sets and immunological signatures, and gene ontology sets are considered as labels (multi-class). Here, to align with other datasets, we use K-means algorithm to cluster the labels for PPI dataset into two groups and use these groups as an indicator to represent the label information for each node in PPI dataset.

\begin{table*}[ht]
	\centering
	\caption{Detailed information of the datasets.}
	\label{Datasets}
	\begin{tabular}{ccccccc}
		\toprule
		Name        & \# Nodes & \# Edges & \# Attributes & \# Labels & Directed? & Weighted?\\
		\midrule
		Cora        & 2708     & 10858    & 1433          & 7         & False     & False\\
		BlogCatalog & 5196     & 171743   & 8189          & 6         & False     & False\\
		Flickr      & 7575     & 239738   & 12047         & 9         & False     & False\\
		PPI         & 56944    & 818716   & 50            & 121       & True     & False\\ 
		\bottomrule
	\end{tabular}
\end{table*}

\subsection{Baseline Methods Setup}
Our experiments evaluate the joint representation through \emph{SANE} on standard supervised learning tasks: label classification on nodes. For this task, we evaluate the performance of \emph{SANE} against the following base line algorithms:

\textbf{Local Linear Embedding (LLE)} \cite{roweis2000nonlinear}: It is a nonlinear dimension reduction algorithm, which seeks a lower-dimensional projection of the data that preserves distances within local neighborhoods.

\textbf{node2vec} \cite{grover2016node2vec}:  It considers truncated random walks generated from the graph as the corpus, and applies word2vec to find the embedding representations for each node.

\textbf{Label informed Attributed Network Embedding(LANE)} \cite{huang2017label}: Based on AANE, it adds the labels information into the procedure of relevant correlation projections.

\textbf{Accelerated Attributed Network Embedding (AANE)} \cite{huang2017accelerated}: It uniformly models the network topology and attributes based on pairwise similarities, and then jointly map them into an identical embedding space via two relevant correlation projections.

\textbf{MultiView} \cite{kumar2011co}: It is a spectral clustering framework that considers the network topology and attributes as two different views, and finds clusters that are consistent across the views co-regularizing the clustering hypotheses.


\subsection{Label classification on nodes}\label{NC}
In label classification on nodes, each node is assigned one label from a finite label set $\mathcal{L}$. 
We test our method and several state-of-the-art networks embedding algorithms on the above datasets by comparing F1 score on the task of label classification on nodes. 
During the training phase, we observe the topology information of the graph, along with the attributes information on the nodes. 
The task is to learn a joint representation, then use this representation, along with a certain fraction of labels information, to predict the label information of the rest.
To be specific, we employ 5-fold cross-validation to report the performance of each algorithm. In each round, we train a logistic regression classifier provided by scikit-learn Python \cite{pedregosa2011scikit} with the default setting on training data (80\% of the whole dataset), and estimate the F1 score on the test data (the rest 20\% of the entire dataset). In all experiments, the joint embedding representation dimension $d$ sets to 96.
Please note that, LANE needs labels as input for the training process as described in \cite{huang2017label}, therefore only for this methods, we expose 80\% of the label information to LANE as the extra input.

Table~\ref{tab:result} reports the F1-scores on \emph{SANE}, along with five baseline algorithms.
From the results, it is evident that the joint representation outperforms the single embedding from topology space only (node2vec) and attribute space only (LLE).
Please note that, LLE, LANE, AANE and MultiView these four methods are based on the matrix factorization which requires the space complexity of $O(N^2)$, these methods suffer from the scalability problem. For example, for PPI dataset, it needs more than $(4 \times 50K)^2 bytes = 40G$ memory space to do the decomposition.
Therefore, we do not report the F1-score on these algorithms on PPI dataset.
On the contrary, \emph{SANE} does not have the above issue due to the use of locality view on both attribute and topological space (see Section~\ref{SC} for a detailed discussion of scalability).

For Cora and BlogCatalog dataset, \emph{SANE} is able to achieve a level of performance equal to or better than matrix factorization. In particular, \emph{SANE} performs neck to neck with LANE on both datasets (with $\approx 1\%$ withdraw). Please note that LANE requires extra labels input during the training process, whereas \emph{SANE} only needs topologies and attributes.
For Flickr dataset, \emph{SANE}'s performance is not as good as Multiview. This is because the performances of two baseline algorithms, LLE and node2vec, are the lowest.
As discussed in Section~\ref{ANE}, \emph{SANE} uses these two algorithms to obtain topology information and attribute information, respectively. However, the performance of the these algorithms perform the worst on Flickr dataset. As a consequence, \emph{SANE} may not be able to obtain the needed information accurately.
However, from the gain of node2vec, we can see that \emph{SANE} achieved the highest increment (up to 71.4\%), this phenomena, on the other hand, proves the validity of the proposed joint representation of \emph{SANE}.
One way to increase the performance is to fine-tune the hyper parameters for LLE and node2vec on Flickr dataset in advance, but this is beyond the scope of this article, so here we omit this procedure.

\begin{table*}[ht]
	\centering
	\caption{Classification performance (F1 score) of different methods on different datasets results.}
	\label{tab:result}
	\begin{tabular}{lccccc}
		\toprule
		& \multicolumn{4}{c}{\textbf{Datasets}} \\
		
		\textbf{Algorithms}  & Cora & BlogCatalog & Flickr & PPI\\
		\midrule
		LLE        & 0.33 & 0.34        & 0.25 & $-$   \\
		node2vec$\triangle$   & 0.79 & 0.65        & 0.42 & 0.69  \\
		LANE$\triangledown$    & 0.84 & 0.92        & 0.88 & $-$   \\
		AANE       & 0.78 & 0.91        & 0.67 & $-$   \\
		MultiView  & 0.82 & 0.83        & \textbf{0.89} & $-$   \\
		\emph{SANE}       & \textbf{0.83} & \textbf{0.91}        & 0.72 & \textbf{0.77}\\
		\midrule
		\emph{SANE} settings ($\lambda, K$)  & (0.8, 127) & (8.0, 112) & (7.2, 100) & (0.8, 81)\\
		\textbf{Gain of node2vec [\%]} & \textbf{5.1} & \textbf{40} & \textbf{71.4} & \textbf{11.6}\\
		\bottomrule
	\end{tabular}
	\begin{tablenotes}[ht]
		\small
		{
			\item \textbf{\emph{Notes}}
			All joint representation dimension fixed to $d=96$.
			$\triangledown$: LANE needs a fraction of label information as well. 
			$\triangle$: We use default node2vec hyper parameters: $p=q=1$.
			$-$： Impractical to calculate due to scalability flaws of the algorithms.
			 
		}
	\end{tablenotes}
\end{table*}

\subsection{Parameter sensitivity}\label{PS}
The \emph{SANE} algorithm involves three important hyper parameters $d$, $\lambda$ and $K$. 
Hence, in this section, we mainly discuss how these three hyper parameters affect the overall performance of \emph{SANE}. 
Except for the parameter being tested, we set other hyper parameters to be fixed, as $p=q=1$, $walk\_length=100$ and $num\_walks=30$.

\begin{figure*}[ht]
	{
		\resizebox{0.95\linewidth}{!}{\includegraphics[width=\linewidth]{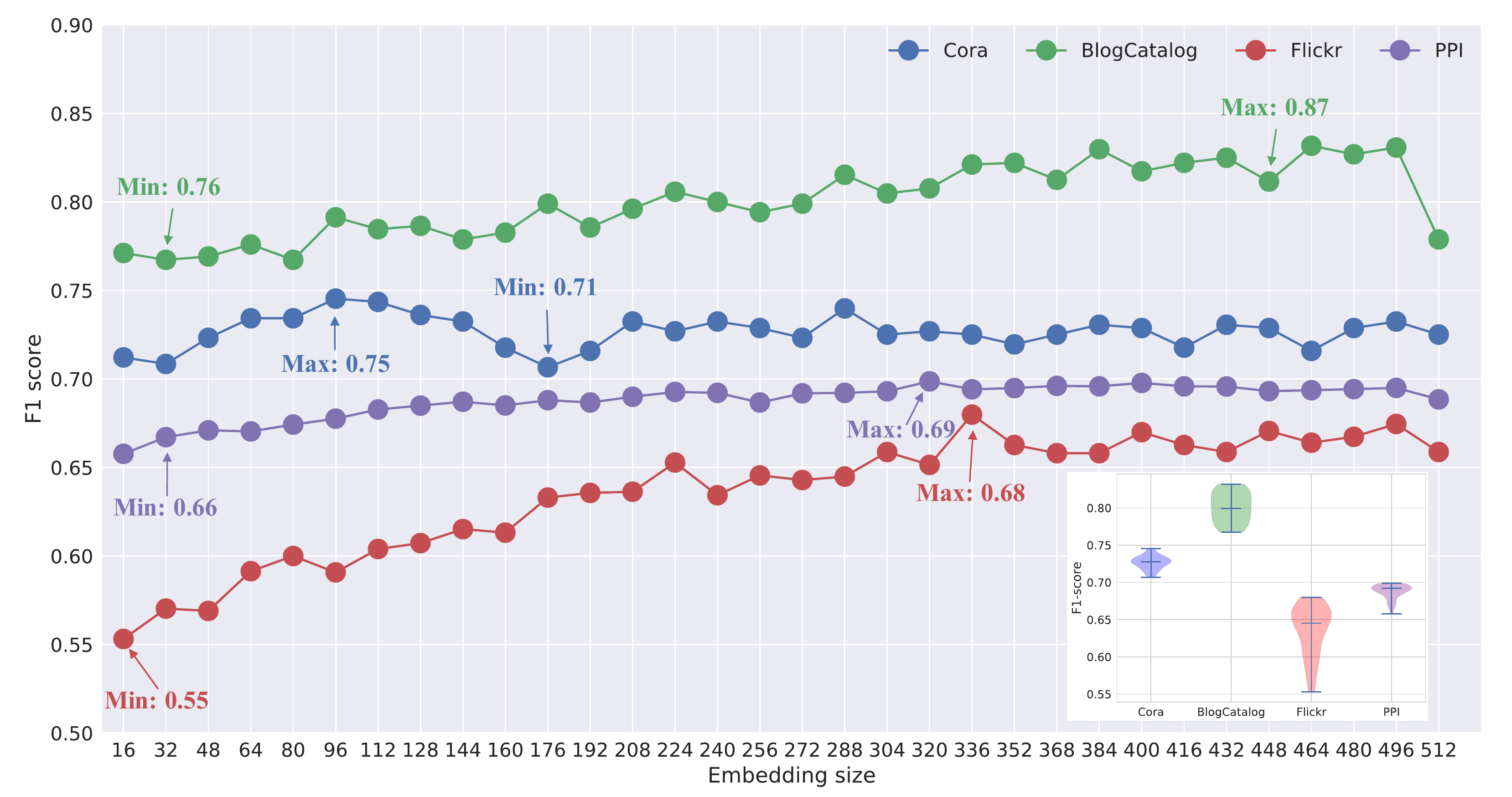}}
		\caption{F1-score on different embedding size. The $x$ axis denotes the embedding size $d$ from 16 to 512, with $step=16$. Whereas the $y$ axis denotes the F1-score. The violin subplot shows the distribution of F1-score for the datasets, respectively.}
		\label{fig:emb}
	}
	
\end{figure*}

\textbf{How $d$ affect the performance}:
The embedding size $d$ indicates the size of hidden vector space that is capable of jointly embedding the nodes, while keeping the relationships from both topology space and attribute space.
By fixing $\lambda=1$ and $K=2$, Figure~\ref{fig:emb} represents the performance with different embedding size $d \in [16,512]$, with $step=16$. 
Here, color ``\emph{blue},'' ``\emph{green},'' ``\emph{red}'' and ``\emph{purple}'' represents the results for different datasets, respectively. For each result, we also mark the minimal and maximal F1-score.
In addition, the violin subplot reports the performance distribution for each dataset, where the lower line, middle line and upper line represents the minimal, median and maximal F1-score respectively.

According to the Figure~\ref{fig:emb}'s results, we can intuitively draw the following conclusion:
The performance of \emph{SANE} on the Cora dataset (``\emph{blue}'') mostly locates near the median, and the different embedding size has a minor effect on the overall performance (only 4\% increment between minimal to maximal), which indicates that the Cora dataset is insensitive to the change of embedding size $d$.
\emph{SANE} on BlogCatalog (``\emph{green}'') shows a uniform distribution across the entire embedding size (shown in violin subplot). 
That is, the performance of \emph{SANE} rises slowing with the increase of the embedding size.
The performance of \emph{SANE} on the Flickr (``\emph{red}'') shows a growth trend below the median and stable trend above the median. In addition, the violin subplot evidently shows that compared with other datasets, the performance of \emph{SANE} is widely distributed (0.55 ~ 0.68), suggesting that the embedding size does affect the performance. Also, we need to note that, when embedding size increases to a certain degree ( $d > 320$,) the performance becomes steady. The explanation for this phenomenon is that, a smaller embedding size is not enough to characterize the properties of Flickr dataset from both topology and attribute space.
The performance for PPI (``\emph{purple}'') can be divided into two points: (1). Similar to Cora, different embedding size has little influence on the performance of the whole (only 3\% increase). However, (2). Like Flickr, the larger the embedding size is, the better \emph{SANE} can capture the essence of the data.

Based on the above analysis, we can conclude: Under the different datasets, \emph{SANE} has different behaviors on different embedding size. So for a practical application, we should first find an embedding of the relationship between the size and performance, before carrying on searching other parameters. 
In fact, for the Flickr data, when $d=336$, $\lambda=8$, $k=81$, the F1-score of \emph{SANE} increases to 0.79 ($\approx7\%$ increment compare to Table~\ref{tab:result}).

\begin{figure}[ht]
	{
		\centering
		\includegraphics[width=\linewidth]{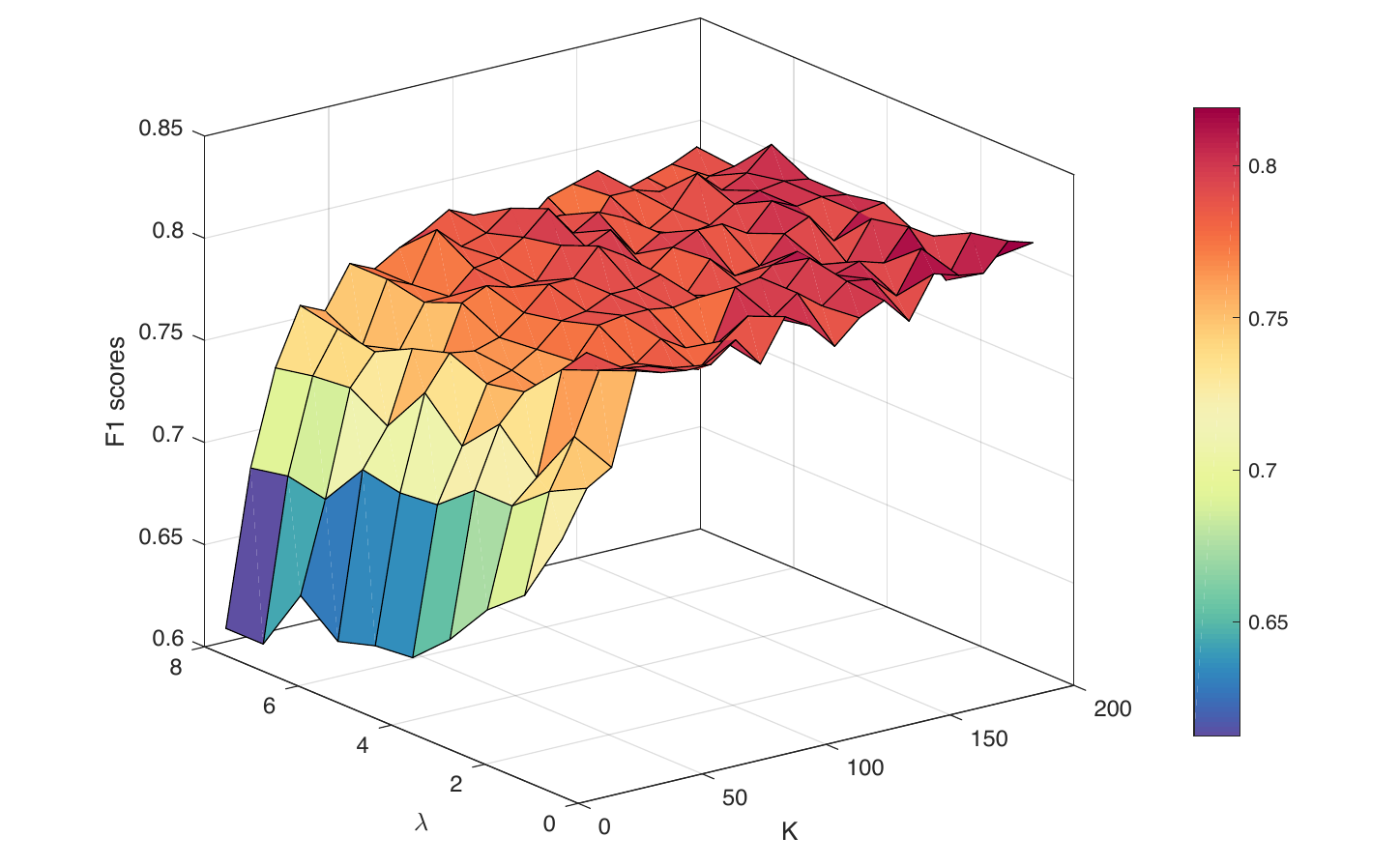}
		\caption{F1-score on different ($\lambda$, $K$) for BlogCatalog dataset. The $\lambda$ axis denotes the weight of LLE from 0.4 to 8.0, $K$ axis denotes the nearest neighbors number of LLE from 1 to 200, and $y$ axis denotes the F1-score.}
		\label{fig:parameter_sensitivity}
	}
\end{figure}

\textbf{How ($\lambda$, $K$) affect the performance}: These two hyper parameters are introduced to learn the joint representation. Here, we use BlogCatalog dataset as an example to demonstrates the performance with $\lambda$ and $K$, on a fixed embedding size $d=96$. Figure~\ref{fig:parameter_sensitivity} reports the result.
It is evident that the influence of ($\lambda$, $K$) on performance shows an upward trend. Among them, $K$ has a more significant impact on performance than $\lambda$. In particular, when $K$ increases from $0$ to $\approx100$, performance grows rapidly; After that, performance turns to grow slowly.
This phenomenon makes sense as the larger the number of neighbors are, the more accurate they can describe the target node. And too many neighbors always contains too many redundant information, which has a minor contribution for learning the joint representation.
It is worth noting that, the impact of $\lambda$ on performance presents an upward trend along with a downward trend (See $K=1$ as an example.) This is because, when $\lambda$ is too large, learning the joint representation will be heavily affected by the information from attribute space, which ignores the features from topology space, thus leads to a decrease in performance.

\subsection{Scalability}\label{SC}
To test for scalability of \emph{SANE}, we learn joint representations for synthetic networks generated from Barabási-Albert preferential attachment model \cite{barabasi1999emergence}. 
We implemented the \emph{SANE} on an Ubuntu 16.10 laptop with an Intel i7-6500U processor (4 cores) and 12GB of 2133MHz RAM， using mixed programming (Python and C++) implementation. 
To be specific,
for ``preprocessing'' part (a.k.a, building corpus from topology and generating neighbors and weights from attributes), we use Python-3.5 as the main implementation language;
for ``learning joint representation'' part, we use C++-11 as the implementation language, g++-6.2 as the compiler.

\begin{figure}[ht]
	\includegraphics[width=\linewidth]{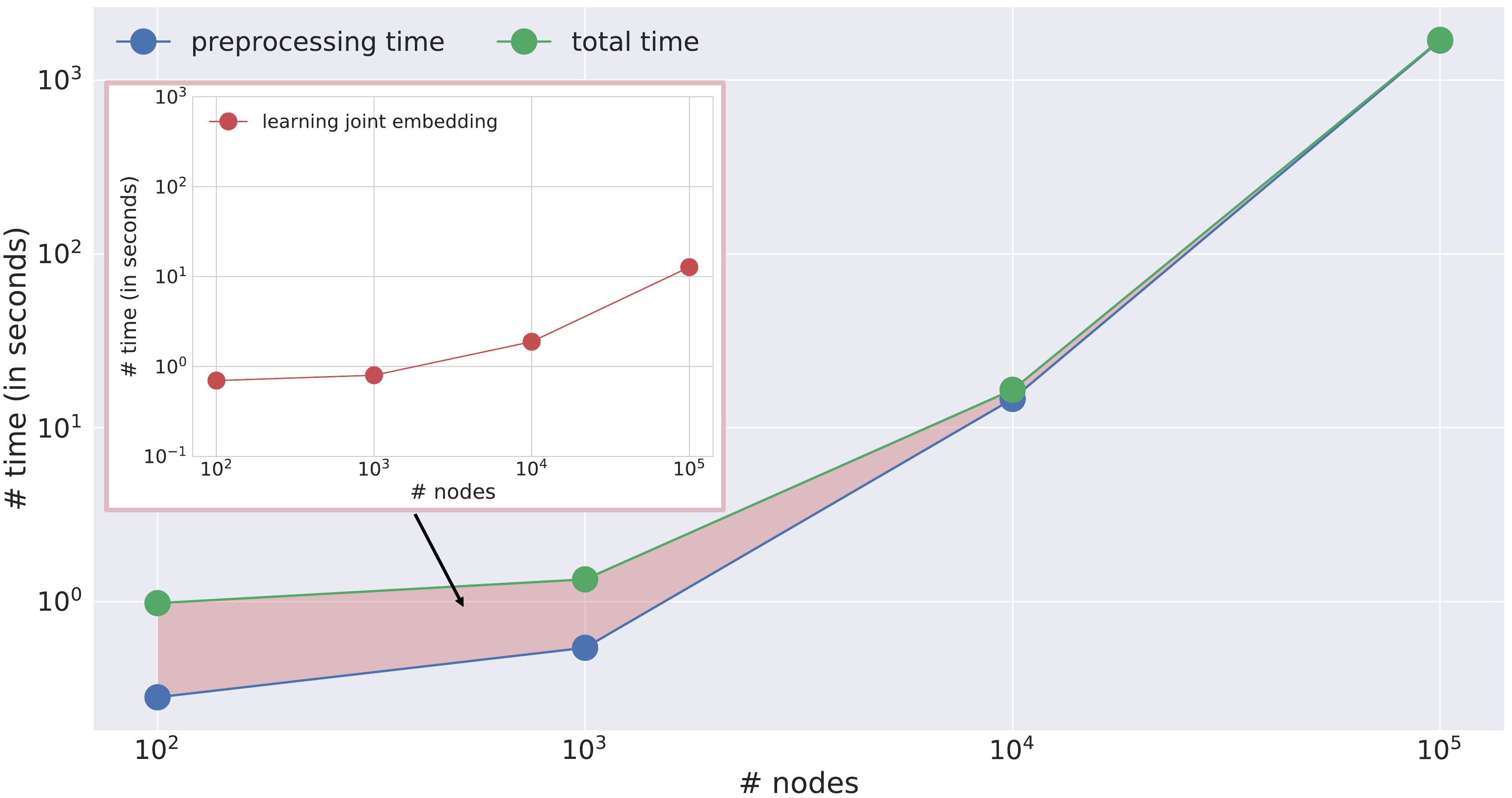}
	\caption{Scalability of proposed method on Barabási-Albert graphs with a fixed attached edges of 3. The subplot shows the scalability of the core step (a.k.a, joint representation) of \emph{SANE}. }
	\label{fig:scalability}
\end{figure}

Figure~\ref{fig:scalability} shows consuming time for the synthetic network with increasing sizes from 100 to 100,000 nodes, with a constant number of edges\footnote{Here we fix the number of attached edges as 3.} to attach from a new node to existing nodes. As we use different implementation languages, we use a subplot to show the scalability on ``learning joint representation.'' On main plot, we demonstrate the scalability on ``preprocessing (blue line),'' and the total algorithm (green line).

Here, we empirically observe that \emph{SANE} scales linearly with increase in the number of nodes generating the joint representations. For 100,000 nodes, \emph{SANE} only takes around 20 minutes, particularly, the joint representation written by C++ only takes $\approx10$ seconds. The optimization phase is made efficient using negative sampling on batches \cite{mikolov2013distributed,ji2016parallelizing}, and asynchronous stochastic gradient decent\cite{recht2011hogwild}.
As the implementation language (C++/Python) and the number of cores is secondary to the overall performance, we believe that the preprocessing time can reduce a lot by optimizing these settings.

\section{Conclusion and Future Work} \label{Conclusion}
Jointly incorporating the network topology and node attributes into network embedding is promising but challenging. Therefore, we propose a novel and scalable framework SANE, which learns unified embedding representations by enforcing the alignment of localized linear mapping from the two spaces. Without the eigen-decomposition of a large matrix, SANE can easily support large-scale attributed network embedding. Several experiments on real-world datasets demonstrate that SANE is indeed an effective and scalable algorithm. 

Future work will focus on: (1) Most attributed networks are evolving, how can we incrementally update embedding representations without relearning the joint representation from scratch? (2) There are always mis-alignments between topology space and attribute space, so how can we selectively enforce the alignment of the localized linear mapping from two different spaces that are truly aligned? (3) Label information sometimes is available for learning embedding representations, so how can we incorporate label information into our current algorithm?

	\bibliographystyle{ACM-Reference-Format}
	\bibliography{sample-bibliography}
	
\end{document}